\documentclass{article}

\usepackage[preprint]{corl_2021} 

\usepackage{tabularx}
\usepackage{graphicx}
\usepackage{booktabs}       
\usepackage{amsfonts}       
\usepackage{amsmath}
\usepackage{amssymb}
\usepackage{nicefrac}       
\usepackage{microtype}      
\usepackage{float}
\usepackage{multirow}

\usepackage{subfig}
\usepackage[capitalise]{cleveref}
\usepackage[font=footnotesize,labelfont=bf]{caption}

\usepackage{amsmath,amsfonts,bm}

\def\Figref#1{Fig.~\ref{#1}}

\def\eqref#1{equation~\ref{#1}}

\def\Tabref#1{Table~\ref{#1}}

\def\Appref#1{Appendix~\ref{#1}}

\def\1{\bm{1}}

\DeclareMathAlphabet{\mathsfit}{\encodingdefault}{\sfdefault}{m}{sl}
\SetMathAlphabet{\mathsfit}{bold}{\encodingdefault}{\sfdefault}{bx}{n}

\title{Playful Interactions for Representation Learning}

\author{
  Sarah Young\\
  UC Berkeley\\
   \And
   Jyothish Pari \\
   NYU \\
   \And
   Pieter Abbeel \\
   UC Berkeley \\
   \And
   Lerrel Pinto \\
   NYU \\

}

\begin{document}

\maketitle
\begin{center}
   {\tt  \small \href{https://sarahisyoung.github.io/play.html}{Project Page}}
\end{center}


\begin{abstract}
One of the key challenges in visual imitation learning is collecting large amounts of expert demonstrations for a given task. While methods for collecting human demonstrations are becoming easier with teleoperation methods and the use of low-cost assistive tools, we often still require 100-1000 demonstrations for every task to learn a visual representation and policy. To address this, we turn to an alternate form of data that does not require task-specific demonstrations -- \emph{play}. Playing is a fundamental method children use to learn a set of skills and behaviors and visual representations in early learning. Importantly, play data is diverse, task-agnostic, and relatively cheap to obtain. In this work, we propose to use playful interactions in a self-supervised manner to learn visual representations for downstream tasks. We collect 2 hours of playful data in 19 diverse environments and use self-predictive learning to extract visual representations. Given these representations, we train policies using imitation learning for two downstream tasks: \emph{Pushing} and \emph{Stacking}. We demonstrate that our visual representations generalize better than standard behavior cloning and can achieve similar performance with only half the number of required demonstrations. Our representations, which are trained from scratch, compare favorably against ImageNet pretrained representations. Finally, we provide an experimental analysis on the effects of different pretraining modes on downstream task learning.


\end{abstract}

\keywords{Self-supervised learning, playful interactions, imitation learning.}

\section{Introduction}

Imitation learning has proven to be a powerful approach to learn complex robotic skills from visual observations~\cite{zhang2018deep, stadie2017third,duan2017one,zhu2018reinforcement}. Recent works have shown how simple approaches like behavior cloning can reliably replicate manipulation behaviors without needing explicit reward feedback~\cite{ho2016generative, kalashnikov2018qt,kaelbling1996reinforcement}. However, such methods are notoriously data hungry, often requiring 100-1000 demonstrations during training. These demonstrations further need to be collected across a wide variety of diverse environments to prevent overfitting to narrow distributions of environments. This paradigm of visual imitation becomes even less practical when we need to learn a multitude of diverse skills for our robots.

But why does visual imitation require such large amounts of data? One hypothesis is that the imitated policy not only needs to learn the desired behavior, but also the appropriate low-dimensional representation for the high-dimensional visual inputs. Hence one path to efficient visual imitation is to reduce the burden of representation learning by using pretrained representation learning models. Using such pretraining both from labelled and unlabelled data is routine in Computer Vision~\cite{ gidaris2018unsupervised,moco2,simclr}. However, in the context of robotics, obtaining reliable pretraining is not straightforward. Standard vision datasets \cite{krizhevsky2012imagenet,cifar10, pascal-voc-2012} contain predominantly outdoor images with various object-centric biases. While, standard robotic datasets \cite{mandlekar2018roboturk,sharma2018multiple} are often lab-specific and contain their own robot-specific biases. In fact, even finetuning from data collected through other tasks on the same robot may not be amenable due to distributional mismatch~\cite{saenko2010adapting, torralba2011unbiased, zhang2019transfer}. This brings us to our central question -- How can we get data that matches the visual distribution of a given robot?



To answer this, we take inspiration from research in human development and look at an alternate form of data: \textit{play}~\cite{piaget2013play, cook2011science, sim2017learning,lillard2013impact,whitebread2017role}. 
From a pure data perspective, playful interactions possess two key qualities. First, it would be cheap to obtain since play is task-agnostic, and it does not need extensive curation or instruction to data collectors. Second, it would be naturally diverse since playful interactions can be easily collected in unstructured environments. But how does one collect and learn from playful interactions for robots?


In this work, we present a framework for representation learning that can scalably collect and learn from playful interactions. First, we use reacher-grabber tools~\cite{song2020grasping} built on top of DemoAT~\cite{young2020visual} to collect play data in the wild. We simply instruct users to “do whatever they want with this tool”. With around two hours of self-guided play, we obtain 30,000 frames of playful interaction data in diverse environments. Equipped with this data, we then use a novel self-supervised learning approach to learn a visual encoder that can extract visual representations. Since play data neither solves a specific task nor operates in a single environment, the obtained visual encoder is task-agnostic and can operate on diverse visual inputs.

To demonstrate the usefulness of representations learned for visual imitation, we probe the visual encoder through downstream task-specific finetuning. In this work we consider two downstream tasks, pushing and stacking. Both tasks come with a small number of expert demonstrations collected on the same reacher-grabber setup, with data taken from \citep{young2020visual}. On both tasks, we report significant improvements in behavior cloning metrics and outperform popular methods such as imitation from scratch~\cite{torabi2018behavioral}, data augmentation based imitation~\cite{young2020visual}, ImageNet based pretraining and multi-task transfer~\cite{parisotto2015actor, devin2017learning}. Against our strongest baseline, ImageNet pretraining, we show that play pretraining achieves up to 27\% better MSE performance during test time. Interestingly, when pretrained on top of ImageNet initialization, we achieve up to 38\% better performance than training from scratch.

\begin{figure}[]
    \centering
    \includegraphics[width=0.99\textwidth]{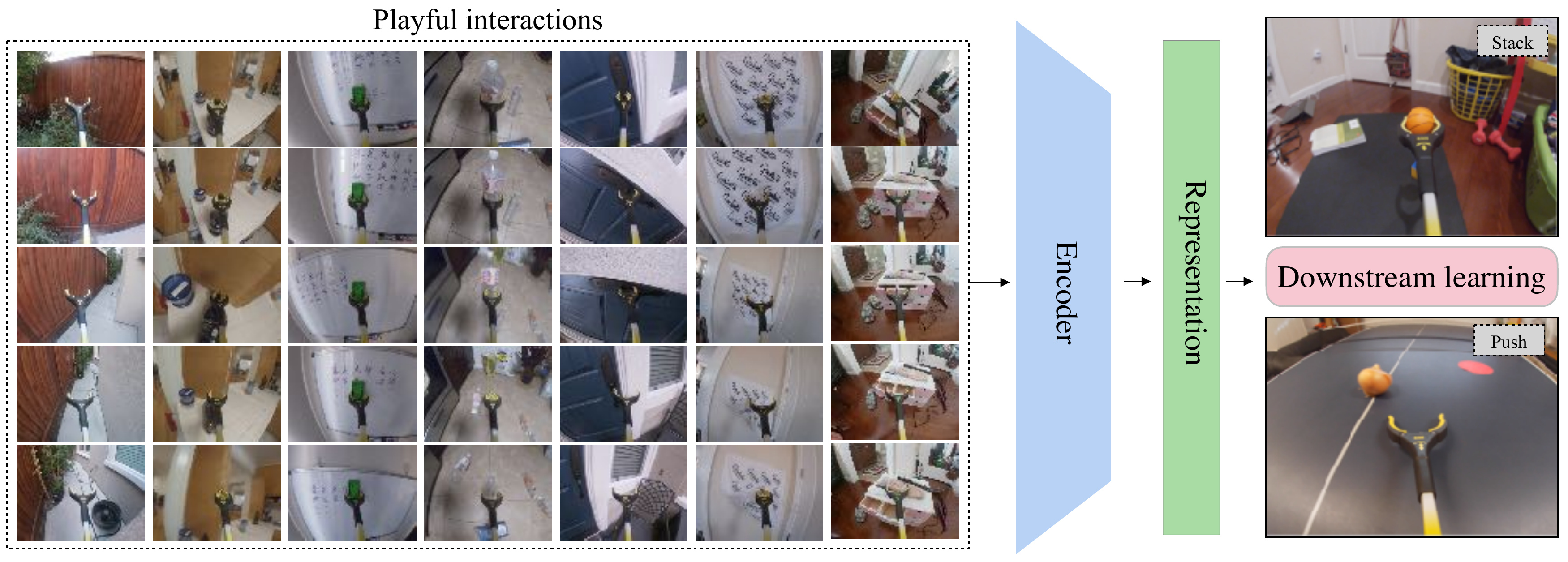}

  \caption{Our method uses around two hours of unlabeled, diverse, and unstructured playful interaction data to learn meaningful representations for downstream manipulation tasks such as Pushing and Stacking.} 
\label{fig:teaser}
\end{figure}

In summary, we present three contributions in this work. First, we propose a framework for collecting playful visual interaction data in the wild. Second, we use self-prediction based representation learning to learn meaningful task-agnostic visual representations. Third, we show that our representations learned on around 2 hours of play can outperform standard imitation-based approaches on two manipulation tasks, pushing and stacking. Although the use of play data has been previously explored in the context of simulated environments \cite{play}, to our knowledge this work is the first that studies the use of this play data in real-world environments. Our playful interaction data, downstream task data, and learned models are publicly released on our website.
 
\section{Related Work}
\subsection{Imitation Learning}
Imitation learning is a common technique used to learn skills and behaviors by observing humans~\cite{piaget2013play, meltzoff1977imitation, meltzoff1983newborn, tomasello1993imitative}. It has been successful in a wide range of robotic tasks such as pushing, stacking, and grasping~\cite{zhang2018deep,DBLP:journals/corr/abs-1802-09564}. Inverse reinforcement learning (IRL)~\cite{abbeel2004apprenticeship} and behavior cloning (BC)~\cite{torabi2018behavioral} are two broad approaches in imitation learning. For a more comprehensive
review of imitation learning, we point the readers to \cite{argall2009survey, hussein2017imitation}.  

Behavior cloning~\cite{torabi2018behavioral} is widely used for its simplicity and ability to generalize to novel scenarios. However, large datasets of expert demonstrations are needed for optimal performance. Demonstrations collected from a different viewpoint or embodiment require complex methods such as transfer learning to overcome the domain gap~\cite{stadie2017third,sermanet2016unsupervised} but are easy to obtain. Large-scale manipulation datasets collected in this manner include \cite{sharma2018multiple, mandlekar2018roboturk}. Recent efforts in eliminating this domain gap include teleoperation methods~\cite{zhang2018deep} and assistive tools \cite{song2020grasping, young2020visual}. Using these assistive tools has shown to be an effective and accessible method for collecting diverse demonstrations at scale. In this work, we adopt the DemoAT \cite{young2020visual} framework to collect expert demonstrations and imitate via behavioral cloning.

\subsection{Learning from Play}
While work in learning from play in robotics has been relatively limited for real robotic tasks, substantial work has been done in simulation. Play-LMP \cite{play} has shown that supervision from teleoperated play data can effectively scale up multi-task learning. In their work, a single goal-conditioned policy is able to perform a variety of user-specified tasks. This demonstrates that playful interactions can learn latent plans capable of task discovery, composition, as well as emergent retrying. In contrast to applying play to planning and multitask scenarios, our work focuses on learning visual representations from imitation. We aim to decrease the amount of task-specific, labeled data needed to learn generalizable policies for manipulation tasks. Furthermore, our definition of play extends to undirected movement beyond object manipulation, incorporating high-dimensional visual observations and operating in the real world.

\subsection{Self-supervised Representation Learning}

Representation learning has long been used in Computer Vision, but interest in this learning technique has recently grown within robotics due to the availability of unlabeled data and its effectiveness in learning tasks. The goal of representation learning is to extract features to improve performance in downstream tasks. The key idea is to exploit information from data without explicit labeling. Unlabeled data is generally first trained on one or more pretext tasks to learn a representation. These tasks can include predicting image rotations and distortions, patches, frame sequence prediction, or instance invariances~\cite{gidaris2018unsupervised,dosovitskiy2015discriminative,doersch2016unsupervised,misra2016shuffle,simclr,moco2,wu2018unsupervised}. The performance of this pretext task is typically discarded, and the pretrained model is used to learn different downstream tasks. Some works have proposed simultaneously training these pretext tasks alongside the main objective~\cite{zhai2019s4l, sun2019unsupervised}. The idea behind pretraining on pretext tasks is that the learned representations have useful structural meanings and are relevant to downstream tasks. A number of works \cite{simclr, moco2, byol, swav, yarats2021reinforcement} have demonstrated state-of-the-art performance with unsupervised representation learning. We follow a BYOL-style~\cite{byol} framework in our experiments since it provided better performance compared to SimCLR~\cite{simclr}, MoCo~\cite{moco2}, and Swav~\cite{swav}.

\subsection{Representation Learning in Robotics}
Learning directly from high dimensional visual inputs is challenging and data inefficient, but contains rich information needed for learning. Representation learning is a popular method which learns low dimensional latent representations from raw images. Several recent works \cite{xie2020learning,botteghi2021low,wulfmeier2021representation} have explored learning latent representations representations from images in the context of robotic tasks. \citet{finn2016deep} combines a deep spatial Autoencoder with RL to learn a state space representation for robotic manipulation tasks. \citet{jonschkowski2015learning} focuses on learning robotic priors by first learning a 2D representation from random actions taken by the robot, followed by a standard reinforcement learning policy. \citet{hoeller2021learning} uses a sequence of images to learn a latent representation for downstream navigation tasks that are trained via an LSTM network. In our work, we combine this idea with learning from play to demonstrate the effectiveness of representations based on real world playful interactions for downstream manipulation tasks.

\section{Approach}

\subsection{Playful Interactions}
We define “playful interactions" as interactions of any kind in a real-world environment using the DemoAT~\cite{young2020visual} framework. We asked four people to collect data, and these users were untrained and given no information about the downstream tasks. The only guideline we gave data collectors was to ``walk around with the reacher-grabber tool and do whatever you want". This includes walking and exploring the space, picking up and placing objects, as well as accidental drops and undirected actions. This style of data is very different from our task-specific data, which only consists of expert, goal-oriented trajectories. Playful interaction data by design is free-form, so there are no categories associated with the data. This kind of unstructured data is useful because it contains exploratory and sub-optimal behaviors that are critical to learning generalizable and robust representations. More importantly, it is much easier to obtain. Since users do not need to be given specific instructions, data collection can be done by any individual, even young children. Furthermore, existing data collected using reacher-grabbers for other purposes can also serve as "playful interaction" data. 

\begin{figure*}[]
  \begin{center}
    \includegraphics[width = 1.01\textwidth]{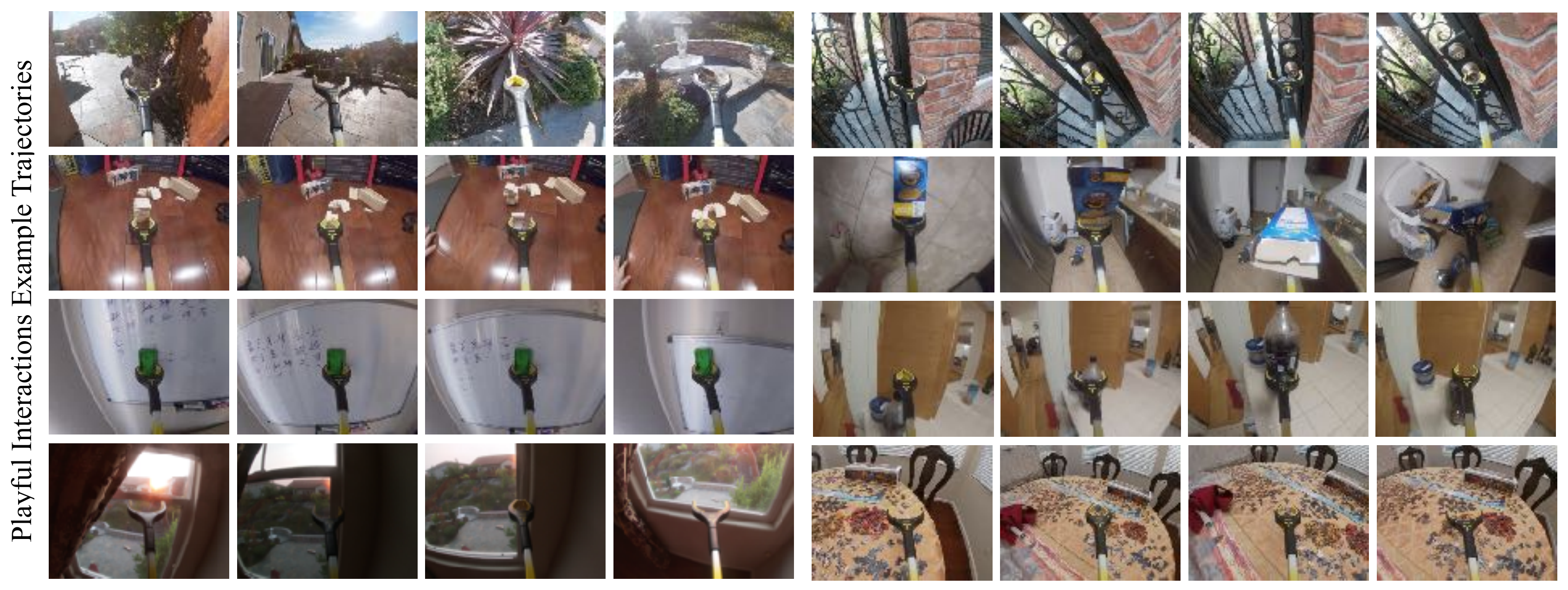}
  \end{center}
  \caption{We show some examples of the playful interaction data. Some trajectories are more free-form and undirected, such as walking around in an open space. Others contain repeated actions, such as erasing a whiteboard or playing Jenga. Many also contain suboptimal behaviors, such as dropping a bottle and knocking it over, and an object sliding out of the gripper because it wasn't grasped correctly. Most importantly, these play trajectories are collected without specific instructions, making it diverse and easy to obtain.}
\label{fig:moreplay}
\end{figure*}

\Figref{fig:teaser} and \Figref{fig:moreplay} display a few examples of playful interaction trajectories. Playful interactions, by nature, will include actions that are very similar to the pushing and stacking task, such as picking up objects. However, there are many more interactions rarely seen in typical demonstration data, including opening and closing doors, erasing whiteboards, playing with blocks, and transitioning from outdoor to indoor settings. We collect 110 minutes of playful interaction data from four different individuals and 19 different locations. In total, we have around 30,000 frames of playful interaction data. Each data collector used the same setup consisting of a reacher grabber stick and a GoPro camera and mount. Our guidelines are flexible enough that data collected from users are quite diverse. For example, demonstrations range from just a few seconds long to up to 18 minutes. Shorter demonstrations tend to be more task-based, while longer demonstrations typically involve many repeated movements and include more undirected interactions such as walking across a room. Our collected playful interaction data will be available on our website. 



\subsection{Learning Visual Representations from Play}
Several prior works \cite{moco2,simclr,oord2018representation,byol} have demonstrated success in pretraining models for downstream visual classification tasks. In our work, we aim to show that pretraining models with playful interaction data is effective for downstream robotics tasks. We choose to use a BYOL~\cite{byol} style framework  to pretrain and learn a visual representation. Unlike the instance-based method used in BYOL, we explore a time-based~\cite{Sermanet2017TCN, purushwalkam2020demystifying} approach to leverage the temporal association available in videos. Instead of augmenting a copy of the same frame, we augment a frame a few timesteps away in the same trajectory. Unlike \cite{Sermanet2017TCN}, however, we do not require paired viewpoints of the same observation. We learn a representation purely from comparing observations from a single viewpoint at different timesteps. We find that a time-based approach is much more effective than the purely instance-based method used in BYOL. 

We train visual encoders $q(\cdot)$ and $k(\cdot)$ for the query and keys respectively and use a momentum-based update for the query encoder. The query encoder $q(\cdot)$ and key encoder $k(\cdot)$ are identical convolutional neural networks. They each take in a single image $I_t \in \mathbb{R}^{3 \times 224 \times 224}$ and output a vector $v$. $I_t$, which is an augmented version of the frame at timestep $t$, is fed into the query encoder, and $I_{t+3}$, which is an augmented version of the frame at timestep $t+3$, is fed into the key encoder. We then feed $v$ into a MLP projection head $h(\cdot)$ and return the latent representation $x_t \in \mathbb{R}^{128}$ for each image. Then, we compute a simple L2 loss between these latent representations. The projection head $h(\cdot)$ is discarded after the self-supervised pretraining phase. 

The play encoder architecture is as follows. Let $Ck$ denote convolutional layers with $k$ filters and $Fk$ denote fully connected layers of size $k$. The base encoder architecture we use for play pretraining is simply the first three convolutional layers of the AlexNet: $C64-C192-C384$, followed by a pooling layer and a MLP projection head of size $F384-F128$. We find that pretraining only the first three convolutional layers rather than four or five layers improves the model's ability to learn and generalize during downstream task evaluation and is key to good performance. In  \Appref{app:layer_ablation}, we provide analysis of pretraining at different layers.

\subsection{Downstream Learning} 

After training on playful interaction data to learn a meaningful representation, we use this representation for downstream manipulation tasks. Unlike other works that utilize self-supervised contrastive pretraining \cite{moco2,simclr,atc,byol}, our network architecture for downstream tasks builds on top of the pretraining encoder and continues to update representation weights.

\noindent \emph{Visual Encoder Architecture:} 
The network architecture for downstream task learning consists of the base encoder used during pretraining followed by two additional convolutional layers and one projection layer: $C64-C192-C384-C256-C256$. During training, weights from every layer are updated during task learning. The encoder architecture is shown in \Appref{app:architecture}.

\noindent \emph{Behavior Cloning:}
We learn a policy using behavior cloning~\cite{pomerleau1989alvinn, ross2011reduction}. Each dataset contains observation-action pairs $D = \{(I_t, a_t)\}$, where $I_t$ is an image and $a_t$ is the action to get from $I_t$ to $I_{t+1}$. Our task-training model takes in an observation image $I_t \in \mathbb{R}^{3 \times 224 \times 224}$ and learns a function $f(I_t, a_t) $ that maps observations $I_t$ to actions $a_t$. Action labels are the relative changes in pose across frames and are provided by the dataset. Thus, we can represent the dataset of expert task-demonstrations as observation-action pairs
$(o_t, \Delta x_t))_{t=0}^T$, where $ x_t \in \mathbb{R}^3 $. 


Our objective is to minimize a combined direction and mean squared error (MSE) loss. MSE is computed on the predicted translation vectors. 



\section{Experiments}
We designed our experiments to address four key questions. First, does self-supervised pretraining with playful interactions capture a diverse set of environments to improve visual imitation? Second, are play representations better than representations learned on ImageNet? Third, would task-specific representations do as well as task-agnostic representations learned from play? Finally, can play representations be combined with other modes of pretraining to get better performance? In this section we assess the performance of our method on two manipulation tasks, pushing and stacking. We will start with describing the foundation of our experiments, the task setup and baselines, and then progress to describing experiments that demonstrate the advantages of pretraining on play. Finally, we dissect the model and convey best practices for pretraining in our ablation study.

\begin{figure*}[]
	\centering
	\includegraphics[width=\textwidth]{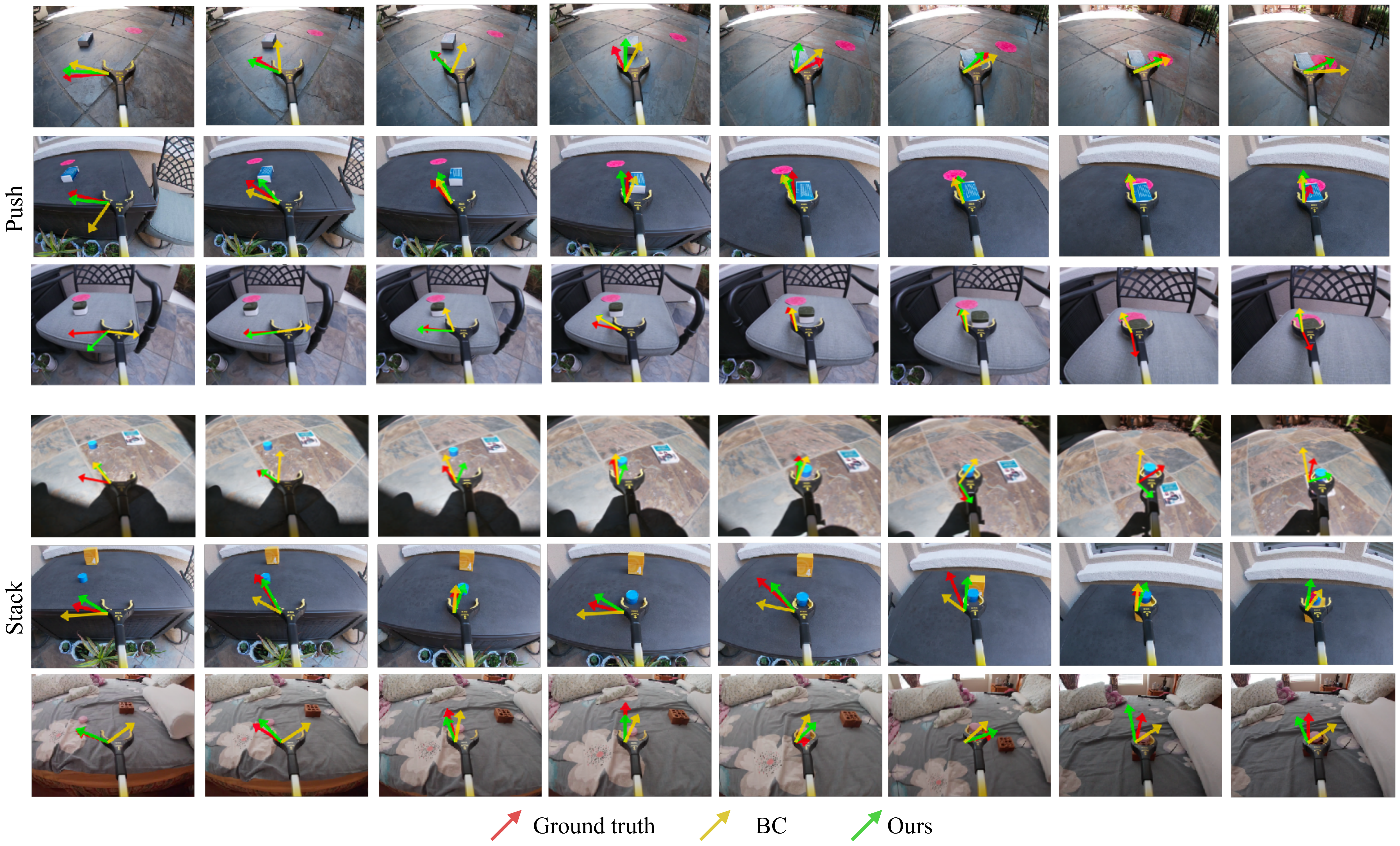}
  \caption{Here we present qualitative results of our experiments for both tasks. Each row represents one trajectory. Each overlaid arrow on the images represents the action predicted by each policy. We only display the predicted translation in the transverse plane of the camera and omit up-down actions in our visualization. We note that because the camera is attached at a forward angle, an arrow pointing downwards (as seen in the the stack trajectories) signifies an upward movement and vice versa for an upwards arrow. We can see that our method consistently predicts actions that are closer to the ground truth action compared to BC.}
\label{fig:qual}
\end{figure*}

\subsection{Downstream Task Setup}
We evaluate our approach on two tasks, pushing and stacking. We use subsets of the 1000 pushing and stacking examples provided in \cite{young2020visual}. 

The goal of the pushing task is to slide an object across a flat surface onto a red circle. The diverse dataset includes demonstrations of around 20 different objects in many diverse scenes, which makes accurately manipulating objects especially challenging. Labels are action deltas between consecutive images, which consist of a translation vector. The prehensile stacking task requires grasping an object and placing it onto another object. To avoid ambiguity during training, the closer object is always stacked above the farther object in the given expert demonstrations. Similar to the pushing task, demonstrations consist of a diverse set of objects and configurations, which significantly increases the difficulty of the task. Test-time demonstrations for each task are collected in completely different environments with new objects. We evaluate our method using MSE on 100 held-out video demonstrations for both tasks to better demonstrate the effectiveness of playful interaction pretraining in the absence of large-scale labeled data. Examples of pushing and stacking tasks are illustrated in \Figref{fig:teaser}. We provide training details in \Appref{app:train_details}.


\subsection{Baselines}
We compare our results to competitive behavior cloning baselines.  \cite{young2020visual} has shown that employing data augmentations such as random jitters, crops, and rotations on top of naive behavior cloning is able to generalize much better to unseen environments during test time. Hence, we use behavior cloning with data augmentations as the base option. Note that every experiment employs the same set of data augmentations. Concretely, our baselines are as follows:
\begin{itemize}
  \item \emph{BC}: a behavior cloning policy trained from scratch, with data augmentations. 
  \item \emph{AE}: a baseline pretrained on playful interaction data via an Autoencoder rather than BYOL. It first learns a representation by minimizing a reconstruction loss, and the learned weights are then used for training downstream tasks.
  \item \emph{VAE}: another generative modeling method used as a representation learning method in several works \cite{vincent2008extracting, play}. Similar to \emph{AE}, this baseline is first pretrained using a VAE before its weights are loaded into the downstream task learning model.
  \item \emph{PLAY}: these baselines are the models we train with playful interaction data.
  \item \emph{BC-OTHER}: an ImageNet initialized baseline pretrained not on playful interaction data, but data used for other tasks, to really see how effective playful interaction data is for visual representation learning. Specifically, we pretrain on the stacking task for the downstream pushing task, and vice versa for the stacking task. 
\end{itemize}

All baselines appended with \emph{-I} are ImageNet pretrained baselines, where we load first a model with weights that have been trained for ImageNet classification tasks, rather than from scratch. We then use this pretrained model to run BC for our downstream manipulation tasks. \emph{PLAY-I} is first loaded with ImageNet pretrained weights, then pretrained on the play dataset, and finally evaluated on downstream tasks. 
Unless explicitly noted, all experiments are evaluated on 100 held-out task-specific demonstrations collected in novel, unseen environments. We include performance of baselines trained with 200 demonstrations in \Appref{app:twoxdata} to show how our method compares to BC-A baselines that train with twice the number of demonstrations.

We find that generative methods such as AE and VAE do not work well. In particular, these baselines, which are first pretrained with playful interaction data and then run on downstream tasks, perform similarly to the BC baseline. We find that both AE and VAE are unable to reach the same level of accuracy as our method during training, and thus does not perform as well on held-out data during test time. We also find that learning with instance discrimination does not improve performance and performs similarly to the BC baseline. This could be due to the fact that BC itself includes data augmentations, so the instance-based pretraining does not give any significant new information.

\begin{table}[]
\caption{Test Mean Squared Error of Playing and Stacking Task (lower is better).}
\label{tab:main_table}
\centering
\begin{tabular}{@{}lllll|lllll@{}}
\toprule
Task  & BC    & AE    & VAE   & PLAY  & BC-I           & AE-I  & VAE-I & PLAY-I         & BC-OTHER \\ \midrule
Push  & 0.095 & 0.101 & 0.084 & \textbf{0.068} & 0.08  & 0.093 & 0.085 & \textbf{0.059} & 0.085    \\
Stack & 0.137 & 0.139 & 0.135 & \textbf{0.129} & 0.126 & 0.138 & 0.137 & \textbf{0.104} & 0.128    \\ \bottomrule
\end{tabular}
\end{table}

\subsection{Does Training on Playful Interactions Lead to Good Representations?}
\label{exp_1}
To test whether self-supervised pretraining with playful interactions can learn a meaningful representation, we first train a model using our collected playful interaction data via BYOL. Then, we load the learned weights into our model to train on the downstream task. We train on 100 trajectories for both the pushing and stacking task. If our playful interactions can learn effective visual representations, we expect that this policy will outperform one where the downstream task is directly trained with BC from scratch. As shown in the first (BC) and fourth (PLAY) columns of \Tabref{tab:main_table}, we see that our play model is able to achieve significantly better results, decreasing MSE from 0.095 to 0.068 in the pushing task and 0.137 to 0.129 in the stacking task. The performance gap is apparent when we visually compare actions between BC and our play model, as shown in \Figref{fig:qual}.

\subsection{How does Pretraining on Playful Interactions Compare to ImageNet Pretraining?}
\label{no_imagenet}
We study whether playful interactions are able to provide enough diversity and information in its learned representation to surpass the performance of BC-I (BC with ImageNet pretraining). The BC-I baseline is trained on significantly more data, but does not leverage playful interaction supervision, so we hypothesize that the baseline likely learns a representation better suited for more vision-based tasks. To test our hypothesis, we first train a randomly initialized model to learn a representation from playful interaction data (PLAY). Using this model, we then learn a BC policy on the pushing and stacking task. The baseline BC-I is trained directly on the tasks with ImageNet-pretrained weights. Our results are shown in the fourth (PLAY) and fifth (BC-I) columns of \Tabref{tab:main_table}. We find that our method is more effective than ImageNet training for both the pushing and stacking stack. We provide qualitative results showing predicted actions on held-out test data in \Figref{fig:qual}. Comparing BC-I and PLAY, we see there is an MSE of $0.08$ to 0.068 respectively in pushing and 0.126 to 0.129 respectively in stacking. We further compare our method to the BC baseline trained on twice the number of demonstrations to evaluate whether our playful interactions can reduce the number of demonstrations needed to achieve good performance, which we discuss in \Appref{app:twoxdata}.

\subsection{Does Pretraining on Task-Specific Data Perform Similarly to Pretraining on Play?}
\label{play_vs_other}
We further investigate whether the exploratory task-agnostic nature of playful interaction data is crucial to the learned representation, or if pretraining on another task in the same action space is able to learn a similarly effective representation. To this end, we compare a model pretrained on playful interaction data (PLAY-I) and a model pretrained on a different task (BC-OTHER). Specifically, we test whether a playful interaction-pretrained model outperforms a stack-pretrained model when trained on the pushing downstream task, and vice versa for the stacking task. We note that stacking and pushing have some structural similarity in actions, and that may improve those results. However, we find that when learning the pushing task, pretraining on the stacking data leads no visible improvement. We hypothesize that the pretraining phase is overfitting to the stacking data, and thus does not learn a generalizable representation. We see similar results for the stacking task in the last two columns of \Tabref{tab:main_table}, where pretraining on a different task-specific dataset does not help the model learn a good visual representation for training other downstream tasks. This further shows the effectiveness and importance of using playful interactions to learn a representation that can be used to efficiently learn downstream tasks.


\subsection{Can Play Pretraining be Combined with ImageNet Pretraining to Learn Better Representations?}
We also evaluate our method combined with state-of-the-art pretrained baselines. In this set of experiments, we demonstrate that by combining our play-pretrained model with ImageNet pretraining (PLAY-I), we are able to achieve even better performance. First, we initialize our model with ImageNet weights before pretraining on playful interaction data.  We then train using BC on downstream tasks. As shown in the fifth and second to last columns of \Tabref{tab:main_table}, PLAY-I performs significantly better than the BC-I baseline. Furthermore, PLAY-I, combined with ImageNet pretraining, outperforms PLAY (fourth column).

\subsection{Ablations}
To further understand the effects of play data and subsequent representation learning, we run a suite of ablations described in detail in \Appref{app:all_ablations}. The most significant of these are highlighted below.

\emph{Comparison with twice the amount of downstream task data: }Our method not only surpasses the BC baseline, but also beats performance of the same BC-I baseline trained on 200 labeled trajectories for the pushing task. Our method, trained on only 100 demonstrations, learns with an MSE of 0.059, which is 14\% better than BC with 200 demonstrations despite training on only half the number of labeled data. Our stacking task performs similarly and we provide experimental results and discuss in more detail in \Appref{app:twoxdata}.

\emph{Effect of Pretraining at Earlier Layers}: We also study the effectiveness of representation learning with play at various layers of our model. We perform ablations over pretraining with playful interaction data until the third, fourth, and fifth convolutional layer of the model. We find that  pretraining representations on fewer initial layers and downstream task training on later layers leads to significantly lower MSE. We provide results and more detailed analysis in \Appref{app:layer_ablation}.

\emph{Amount of Play}: In addition, we analyze the increase of performance over the amount of play data. We find that there is a diminishing improvement in MSE when adding more play data and provide more detail in \Appref{app:play_amount}.


\subsection{Connecting to Real Robot Results}

The experimental results in this work are limited to offline MSE evaluations. However, to highlight our MSE evaluation results and contextualize it with real-robot evaluations, we can roughly base our results on \citet{young2020visual}. They show that a MSE of 0.028 corresponds to a 87.5\% success rate for the pushing task and an MSE of 0.06 corresponds to 62.5\% success rate for the stacking task on the real robot. Experimental results in our work have higher MSE since we are operating in the few-shot setting and hence use only a tenth of the pushing and stacking training data used in \cite{young2020visual}. In the context of the experiments in this work, the BC baselines achieve a MSE of 0.08 and 0.126 for the pushing and stacking task respectively (\Tabref{tab:main_table}), which both correspond to not being able to complete either task. The best performing models trained with our method achieve a MSE of 0.059 and 0.104 for the pushing and stacking task respectively, which roughly correlate to successfully solving the task around 60\% for pushing and 29\% for stacking. 


\section{Conclusion}
\label{sec:conclusion}
We have presented an approach for learning downstream manipulation tasks via self-supervised pretraining on easy-to-obtain playful interaction data. Our method improves the generalizability of imitation learning  baselines beyond simple data augmentations and provides significant improvements to current baselines. We demonstrate that our pretraining method can achieve comparable results to behavior cloning baselines using just half of the labeled task data. The success of our technique on simple behavior cloning opens up many exciting avenues for further work to incorporate play into more complex algorithms.



\acknowledgments{
We thank Mandi, Andrew, and Daniel for helping us collect playful interaction data. We also gratefully acknowledge the support from The Open Philanthropy Project, Berkeley DeepDrive, Honda Research Institute and ONR.
}


\bibliography{references}  

\clearpage 

\appendix

\section{Play Encoder Architecture}
\label{app:architecture}
\begin{figure*}[h]
  \begin{center}
    \includegraphics[width = \textwidth]{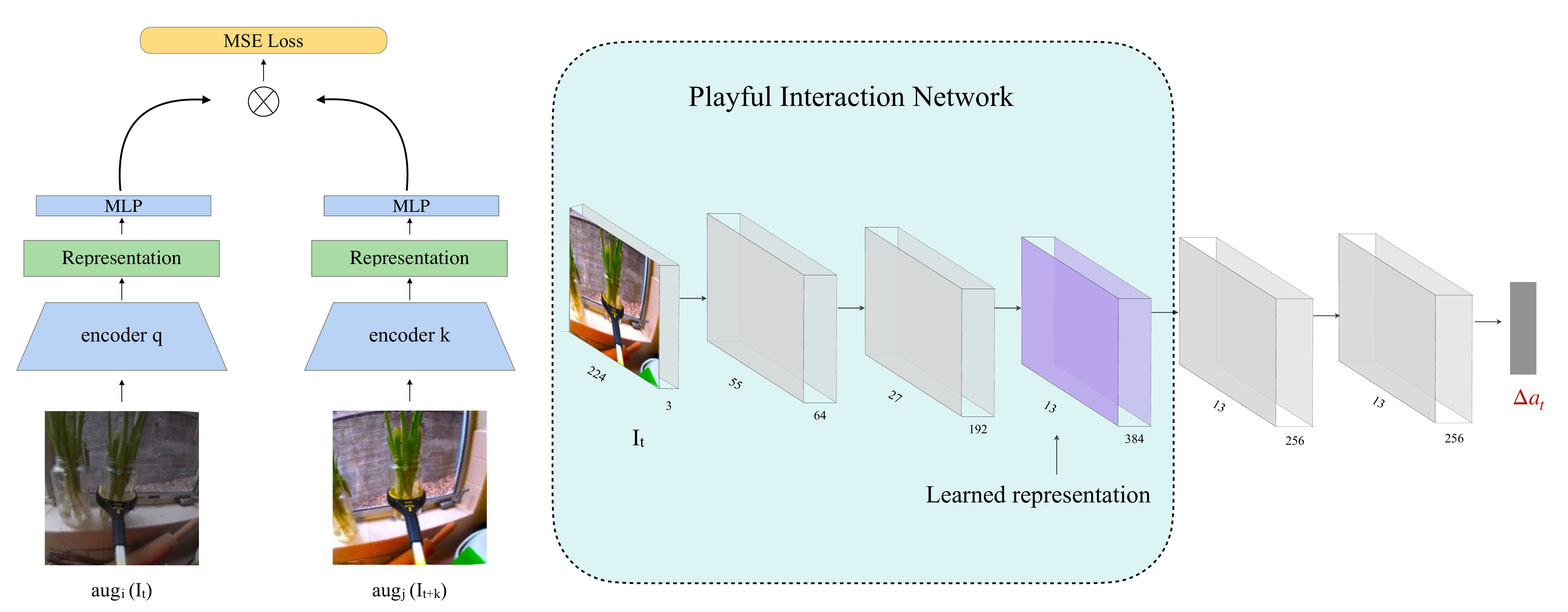}
  \end{center}
  \caption{The setup of our approach consists of a self-supervised pretraining phase and a downstream imitation learning phase. (a) shows the first phase, where a representation is learned via time-based self-supervised learning. Each pair of images is fed through an encoder and projected to an MLP layer before computing similarity. (b) illustrates the encoder architecture for both phases. The purple layer is the representation we optimize during pretraining. The gray region is the encoder architecture used during playful interaction pretraining. The second phase of training (downstream task learning) updates all layers shown and outputs a predicted action. Note that we leave out the projection MLP layers in this image for simplicity. }
\label{fig:arch}
\end{figure*}

The setup of our approach consists of a self-supervised pretraining phase and a downstream imitation learning phase. Here, we show how we train via a time-based BYOL method with playful interactions. In \Figref{fig:arch}, the intermediate layer highlighted in purple is the representation learned during pretraining, and the following two convolutional layers are trained only during task learning. The last layer is a projection to the predicted action.

\section{Training Details }
\label{app:train_details}
We pretrain with 2 hours of playful interaction data, which is around 30,000 frames. The encoders are pretrained for 4,500 gradient steps with a batch size of 64. Then, the weights are loaded into the model for training downstream tasks, and trained for around 1,000 gradient steps. In both phases, we use a batch size of 64. During downstream tasks, we train on roughly 1400 images.


\section{Additional Ablation Studies }
\label{app:all_ablations}

\subsection{Comparison with More Downstream Task Data }
\label{app:twoxdata}
For both the pushing and stacking downstream task, we evaluate our method on 100 trajectories. We have shown that our playful interaction pretraining method is able to significantly improve upon baselines using the same number of expert pushing and stacking demonstrations. However, we also want to evaluate how our method performs compared to using more expert data. To this end, we test the baseline BC-I model on 200 trajectories and compare that to our playful interaction model (PLAY) trained on only 100 trajectories.

\begin{table}[h]
\centering
\caption{Comparison of Amount of Expert Demonstration Data}
\label{tab:extra-table}
\begin{tabular}{@{}llll@{}}
\toprule
Task                     & \multicolumn{2}{c}{BC-I} & PLAY-I \\ \midrule
\# Expert demonstrations & 100         & 200        & 100     \\
Push                     & 0.08        & 0.069      & 0.059   \\
Stack                    & 0.126       & 0.099      & 0.104   \\ \bottomrule
\end{tabular}
\end{table}

 The pushing task trained with PLAY achieves an MSE of 0.059, which is 14\% better than BC with 200 demonstrations despite training on only half the number of labeled data. In the stacking task, our method (0.104) is able to surpass BC performance (0.126) and nearly match BC trained on 200 demonstrations (0.099). We note that in the stacking task, although our method is only able to nearly match performance of using twice the amount of data, it is still a significant improvement upon the baseline. Results are shown in \Tabref{tab:extra-table}.

\begin{figure}[h]
  \begin{center}
    \includegraphics[width = 0.6\textwidth]{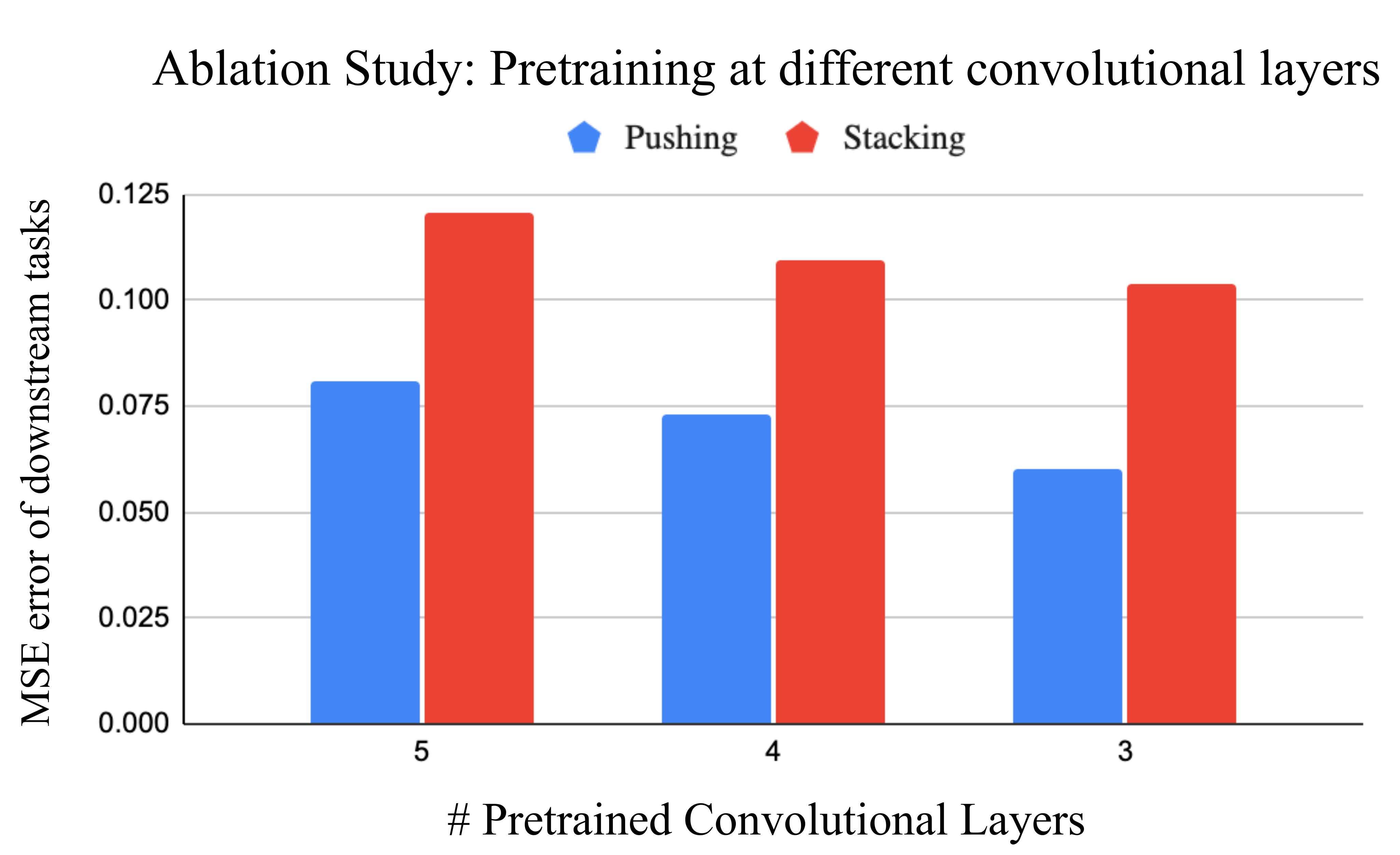}
  \end{center}
  \caption{Learning comparisons between pretraining until different layers. We perform ablations over pretraining with playful interaction data until the third, fourth, and fifth convolutional layer of the model. Note that during downstream task training, we train on the full architecture regardless of which layer we pretrained until. The y-axis is MSE error of predicted and ground truth actions in log scale. We find that pretraining fewer layers significantly improves downstream task performance. We hypothesize that this happens because during pretraining, since we are not optimizing for the same objective as during downstream task learning, the model tends to overfit when trained on more layers.}
\label{fig:layers_graph}
\end{figure}

\begin{figure}[h]
  \begin{center}
    \includegraphics[width = 0.49\textwidth]{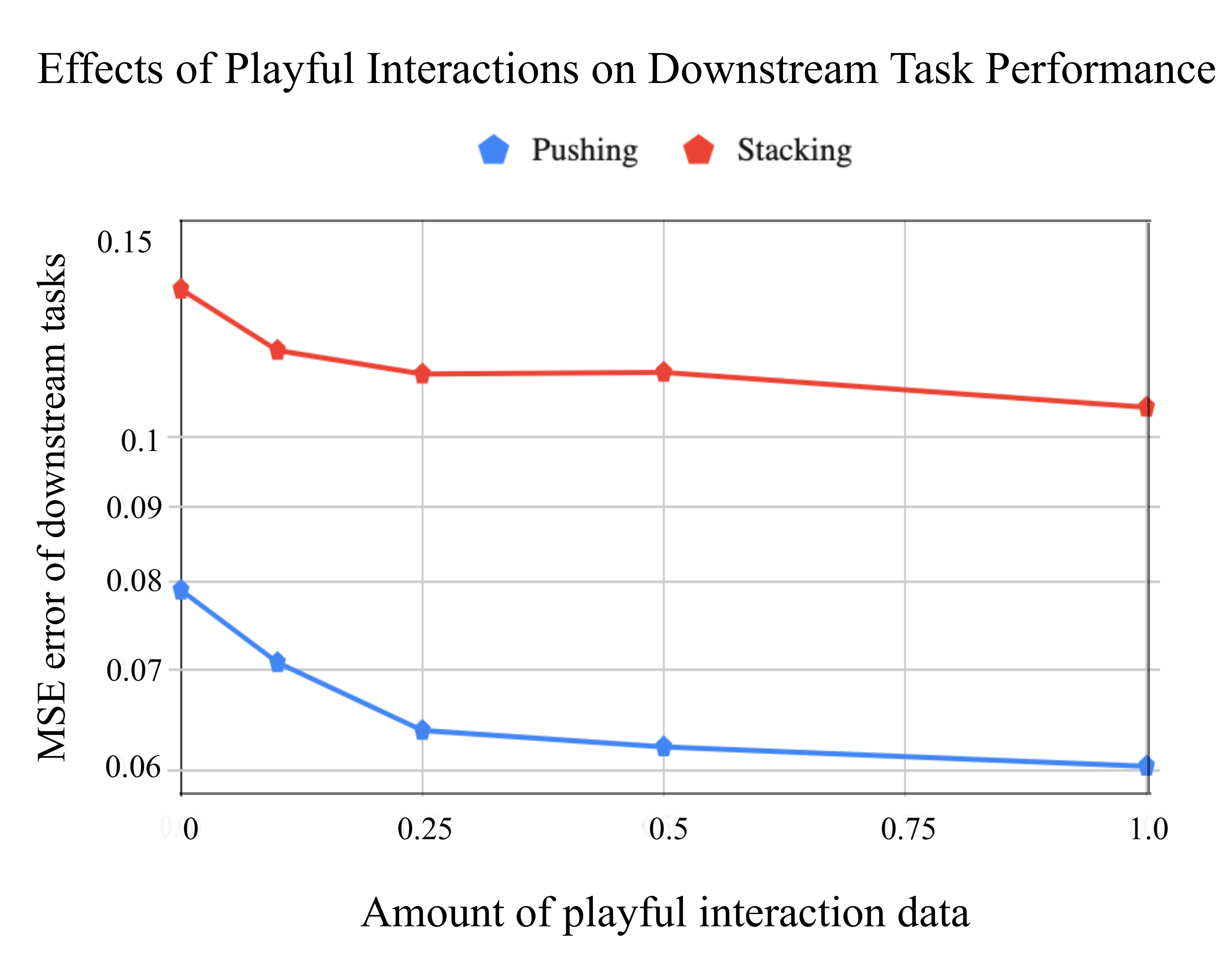}
  \end{center}
  \caption{Learning comparisons between different amounts of play data. Note that the y-axis is MSE error of predicted and ground truth actions in log scale. We see that more play data greatly improves performance on downstream tasks, and diminishing improvements suggest that large-scale play data is needed for improved accuracy.}
\label{fig:amount_chart}
\end{figure}

\subsection{Effect of Pretraining at Earlier Layers} 
\label{app:layer_ablation}
We experiment with representation learning at earlier layers of the base encoder. Specifically, we study the effects of pretraining a representation at the third, fourth, and last convolutional layers shown in \Figref{fig:arch}. We find that representation learning at earlier layers and updating all weights during task learning is crucial to learning a good policy. When learning a representation at the final convolutional layer and fine-tuning only on the last linear layer(s), the policy is unable to learn enough during fine-tuning to successfully complete the task, and performs worse than BC-I (0.093). This suggests that the task-agnostic learned representation itself still needs signal from task-training. If we update all the weights during task learning, however, the model is unable to generalize well and achieves performance (0.081) similar to that of BC-I. Thus, learning a representation at an earlier convolutional layer and updating all weights enables sufficient learning during imitation while preserving the ability to generalize to test time scenarios. \Figref{fig:layers_graph} shows the ablation study comparing performance at different layers of pretraining for both downstream tasks.

\subsection{Amount of Play Data }
\label{app:play_amount}
We study the effects of the amount of play data on downstream task performance. We have 110 minutes of playful interaction videos split into around 30,000 frames and evaluate how important this data is to task learning. \Figref{fig:amount_chart} shows the performance of the pushing and stacking task when pretrained on the following fractions of play data: 1, 10, 25, 50, and 100\%. We note that as the amount of playful interaction data we use increases, the MSE error decreases significantly, especially when the amount of play data is low. Training with just 25\% of our play data shows a 13\% increase in performance when compared to training with 1\% of play data. However, in both the pushing and stacking task, the improvements start to diminish as we approach two hours of data, suggesting that large-scale playful interaction data is needed for further performance gains.

\end{document}